 \documentclass[pmlr,twocolumn,10pt]{jmlr} 

\usepackage{booktabs}
\usepackage{siunitx}

\usepackage[switch]{lineno}

\newcommand{\equal}[1]{{\hypersetup{linkcolor=black}\thanks{#1}}}

\theorembodyfont{\upshape}
\theoremheaderfont{\scshape}
\theorempostheader{:}
\theoremsep{\newline}

\jmlrvolume{LEAVE UNSET}
\jmlryear{2024}
\jmlrsubmitted{LEAVE UNSET}
\jmlrpublished{LEAVE UNSET}
\jmlrworkshop{Machine Learning for Health (ML4H) 2024} 

\title[Edge-AI for Rural Fetal Monitoring]{Edge AI for Real-time Fetal Assessment in Rural Guatemala}

\author{
\Name{Nasim Katebi}\thanks{Emory University, Atlanta, GA, USA.}\thanks{Wuqu' Kawoq | Maya Health Alliance, Guatemala.}, 
\Name{Mohammad Ahmad}\footnotemark[1], 
\Name{Mohsen Motie-Shirazi}\footnotemark[1], 
\Name{Daniel Phan}\footnotemark[1], 
\Name{Ellen Kolesnikova}\thanks{Decatur High School, Atlanta, GA, USA.}, 
\Name{Sepideh Nikookar}\footnotemark[1],
\Name{Alireza~Rafiei}\footnotemark[1],
\Name{Murali~K.~Korikana}\footnotemark[1],
\Name{Rachel Hall-Clifford}\footnotemark[1],
\Name{Esteban~Castro}\footnotemark[2],
\Name{Rosibely~Sut}\footnotemark[2],
\Name{Enma~Coyote}\footnotemark[2],
\Name{Anahi~Venzor~Strader}\footnotemark[2],
\Name{Edlyn~Ramos}\footnotemark[2],
\Name{Peter Rohloff}\footnotemark[2],
\Name{Reza Sameni}\footnotemark[1]\thanks{Georgia Institute of Technology, Atlanta, GA, USA.} \equal{Joint senior authors}, 
\Name{Gari D. Clifford}\footnotemark[1]\footnotemark[4] \footnotemark[5]
}

\begin{document}
\maketitle





\section{Introduction}
\label{sec:intro}
Perinatal complications, defined as conditions that arise during pregnancy, childbirth, and the immediate postpartum period, represent a significant burden on maternal and neonatal health worldwide \citep{who_maternal_health}. Factors contributing to these disparities include limited access to quality healthcare, socioeconomic inequalities, and variations in healthcare infrastructure. Addressing these issues is crucial for improving health outcomes for mothers and newborns, particularly in underserved communities. To mitigate these challenges, we have developed an AI-enabled smartphone application designed to provide decision support at the point-of-care. This tool aims to enhance health monitoring during pregnancy by leveraging machine learning (ML) techniques. The intended use of this application is to assist midwives during routine home visits by offering real-time analysis and providing feedback based on collected data. The application integrates TensorFlow Lite (TFLite) and other Python-based algorithms within a Kotlin framework to process data in real-time. It is designed for use in low-resource settings, where traditional healthcare infrastructure may be lacking. The intended patient population includes pregnant women and new mothers in underserved areas and the developed system was piloted in rural Guatemala. This ML-based solution addresses the critical need for accessible and quality perinatal care by empowering healthcare providers with decision support tools to improve maternal and neonatal health outcomes.

\section{Method}
\label{sec:method}
Over the past decade, we have designed and deployed an affordable perinatal screening system that utilizes low-cost smartphones to collect essential data during routine home visits including Doppler-based fetal cardiac activity, maternal blood pressure (BP) and symptoms indicative of potential complications \citep{Martinez2018, 10.1145/3209811.3209815}. 

The ML models were developed using data captured by midwives in rural Guatemala \cite{Martinez2018}. These midwives collected one-dimensional Doppler ultrasound (1D-DUS) and pictures captured from BP monitors from pregnant women carrying singletons in their second and third trimesters during routine screenings.
\subsection{ML Algorithms}
    {\bf 2.1.1: Doppler Quality Classification:} We developed a deep-learning algorithm that integrates convolutional and recurrent neural networks with an attention mechanism to classify the quality of 3.75-second segments of 1-D Doppler signals into five categories: `Good,' `Poor,' `Interference,' `Talking,' and `Silent' \citep{motie2023point}. This enables midwives to assess the data quality in real-time, allowing them to make informed decisions, such as adjusting the position of the Doppler or improving connections, to capture higher-quality signals.
    
    \noindent {\bf 2.1.2 Heart Rate Estimation}: We introduced a heart rate estimation model based on autocorrelation function for fetal heart rate (FHR) estimation from 1D-DUS signals \citep{valderrama2019open}. This method identifies pseudo-periodic 1D-DUS peaks corresponding to fetal heartbeats by measuring the similarity of a signal to itself over varying time delays, facilitating the detection of the dominant frequency associated with the heart rate.
    
    \noindent {\bf 2.1.3 Fetal Development Assessment:} We designed a hierarchical deep sequence learning model including both convolutional and recurrent networks with an attention mechanism to learn the normative dynamics of fetal cardiac activity in different stages of development and estimate gestational age (GA)\citep{katebi2023hierarchical}.  
    
    \noindent {\bf 2.1.4 BP Image Transcription:} The BP image transcription model operates in two steps: BP monitor LCD localization using the You Only Look Once (YOLO) object detection model \citep{sheppard2020self} to identify the location of the LCD, and digit recognition using a convolutional neural network-based model to recognize and transcribe the sequence of digits displayed on the screen \citep{katebi2023automated, kulkarni2021cnn}.


\subsection{Deployment Pipeline}
We deployed the ML algorithms on a Google Pixel 6a with SDK version 33 (minimum SDK version 28). The Pixel’s 6 GB RAM, Google Tensor chip, and 128 GB storage enabled inference for all models and data processing. The models were deployed using Chaquopy, a Python SDK for Android that enables Python 3.8 integration. The models were converted to a more suitable format for edge devices, TFLite, and the \texttt{tflite\_runtime} library was used for inference. Data pre-processing was done using python scripts running directly within the Android application. To enable real-time analysis of 1D-DUS for quality assessment and FHR estimation, data was processed every second using a ring buffer. This approach allowed for continuous and efficient handling of incoming data stream.
\subsection{Experiments}
To assess the performance of the deployed ML algorithms, we conducted experiments comparing performance on a MacBook Pro (M1 silicon CPU, 16GB RAM, 512GB SSD) using Python, and a Pixel 6a running a Kotlin application to assess consistency and effectiveness in real-world application.

For the quality classification, 3.75-second segments of 1D-DUS recordings from various quality classes were used as test inputs. The classification outputs were then compared across both platforms. Following this, the FHR estimation model was tested on good quality, 3.75-second segments of 1D-DUS recordings, assessing app performance by mean absolute error (MAE) and standard deviation (SD) of error. The GA estimation model was evaluated using recordings, each containing ten concatenated 3.75-second segments of good quality 1D-DUS data, with MAE and SD of error calculated on each platform. Lastly, the BP transcription model was assessed by comparing transcribed systolic and diastolic BP values, with MAE and SD of error calculated for accuracy.


\section{Results}
\label{sec:result}
The app, currently used by 39 midwives and supporting approximately 60 births per month (around 700 women at any given time), was tested using the ML algorithms developed in our previous work. For the quality classification model, we tested a total of 250 segments from the Guatemala dataset, comprising 110 segments labeled as “Good”, 78 as “Poor”, 11 as “Interference”, 17 as “Talking”, and 34 as “Silent”. The classification model deployed on the Pixel 6a app accurately classified all segments, with the exception of one labeled “Interference,” which was misclassified as “Poor.” This misclassification is not expected to significantly affect downstream processes. The FHR model was subsequently tested on 100 segments of 1D-DUS recordings, with mean and SD FHR values of 139.68 and 13.03 beats per minute (BPM), respectively. The MAE between the FHR estimates on the two devices was $0.044 \pm 0.075 $ BPM. The GA estimation model was evaluated on 137 recordings, yielding a mean and SD GA of 34.17  and 4.78 weeks. The MAE between the two devices for GA estimation was $(6.3 \pm 4.9) \times 10^{-6}$ weeks.
Lastly, the deployed BP transcription model was evaluated using images with mean and SD values for systolic and diastolic BP of $104.38 \pm 12.09$ and $65.35 \pm 9.98$ mmHg, respectively. The deployed model on the app demonstrated low error with MAE of $1.15 \pm 8.7$ mmHg for systolic BP and $0.7 \pm 7.3$ mmHg for diastolic BP.

\section{Discussion}
\label{sec:result}
{\bf The key challenges} in this work were included implementing a complex set of machine learning algorithms that worked in a looped buffer, executing on real-time data. The quality and heart rate algorithms had to be optimized to run every second.
We also had to optimize the number of windows to perform gestational age estimation, finding that there was no benefit beyond 10 windows. Further, we found that Doppler-specific and BP device-specific models outperformed generic models. {\bf The key lesson learned} was that co-designing the tool with midwives helped tailoring it to their needs and improve usability. This had a direct impact, through improved midwife engagement, reducing time per visit, and improving diagnostic performance.
{\bf This is highly significant}, with the potential to reduce delay to treatment, and further improve outcomes in this vulnerable and low-resource population. 
\section{Acknowledgement}
This work was supported by a Google.org AI for the Global Goals Impact Challenge Award, the National Institutes of Health, the Fogarty International Center and the Eunice Kennedy Shriver National Institute of Child Health and Human Development, under grant \# 1R21HD084114 and 1R01HD110480. G. D. Clifford is also partially supported by the National Center for Advancing Translational Sciences of the National Institutes of Health under Award \# UL1TR002378. The content is solely the responsibility of the authors and does not necessarily represent the official views of the National Institutes of Health. N. Katebi is funded by a PREHS-SEED award grant \# K12ESO33593.

\bibliography{References}
\end{document}